\documentclass[journal,twoside,web]{ieeecolor}
\usepackage{color}
\usepackage{xcolor}
\usepackage{jsen}
\usepackage{cite}
\usepackage{amsmath,amssymb,amsfonts}
\usepackage{algorithmic}
\usepackage{graphicx}
\usepackage{textcomp}
\usepackage{wrapfig}
\usepackage{stfloats}
\usepackage{subfigure}
\usepackage{booktabs}
\usepackage{multirow}

\def\BibTeX{{\rm B\kern-.05em{\sc i\kern-.025em b}\kern-.08em
    T\kern-.1667em\lower.7ex\hbox{E}\kern-.125emX}}

\begin{document}

\definecolor{abstractbg}{rgb}{0.89804,0.94510,0.83137}
\setlength{\fboxrule}{0pt}
\setlength{\fboxsep}{0pt}

\title{CZU-MHAD: A multimodal dataset for human action recognition utilizing a depth camera and 10 wearable inertial sensors}


\author{Xin Chao, Zhenjie Hou, Yujian Mo
\thanks{Xin Chao is with Jiangsu Digital Twin Technology Engineering Research Center for Key Equipment of
Petrochemical Process, Changzhou University, Changzhou 213000, Jiangsu, China (e-mail: chaoxin941203@163.com). }
\thanks{Zhenjie Hou (corresponding author) is with Jiangsu Digital Twin Technology Engineering Research Center for Key Equipment of
Petrochemical Process, Changzhou University, Changzhou 213000, Jiangsu, China (e-mail: houzj@cczu.edu.cn).}
\thanks{Yujian Mo is with the college of Electronic and Information engineering, 
Tongji University, Shanghai 200000, China (e-mail: yujmo@tongji.edu.cn).}}

\IEEEtitleabstractindextext{%
\fcolorbox{abstractbg}{abstractbg}{%
\begin{minipage}{\textwidth}%
    \begin{wrapfigure}[12]{r}{3in}%
        \includegraphics[width=2.8in]{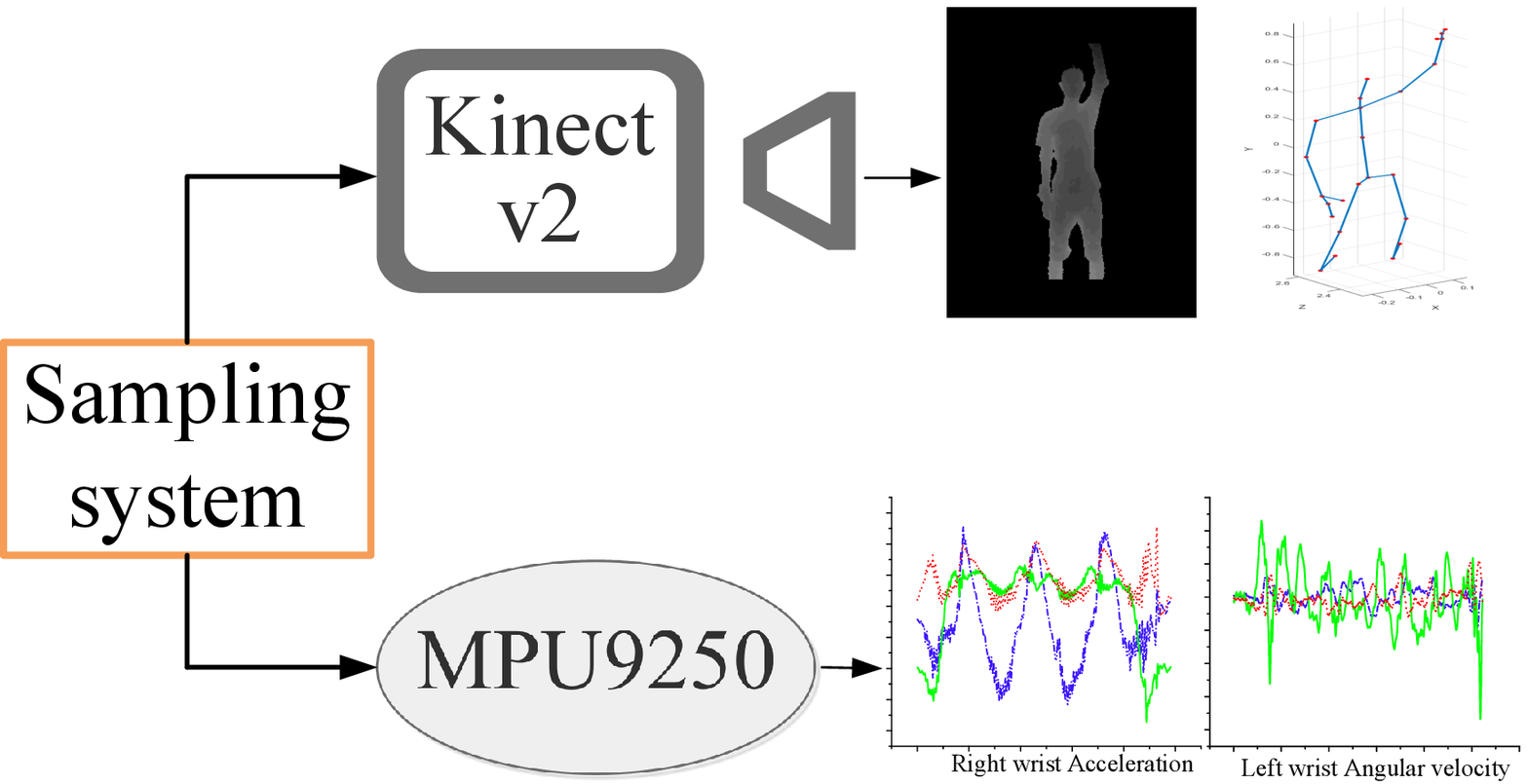}%
    \end{wrapfigure}%
\begin{abstract}
Human action recognition has been widely used in many fields of life, 
and many human action datasets have been published at the same time. 
However, most of the multi-modal databases have some shortcomings in the layout and number of sensors, 
which cannot fully represent the action features. 
Regarding the problems, this paper proposes a freely available dataset, named CZU-MHAD (Changzhou University: a comprehensive multi-modal human action dataset). 
It consists of 22 actions and three modals temporal synchronized data. 
These modals include depth videos and skeleton positions from a kinect v2 camera, and inertial signals from 10 wearable sensors.
Compared with single modal sensors, multi-modal sensors can collect different modal data, 
so the use of multi-modal sensors can describe actions more accurately. 
Moreover, CZU-MHAD obtains the 3-axis acceleration and 3-axis angular velocity of 10 main motion joints by binding inertial sensors to them, and these data were captured at the same time. 
Experimental results are provided to show that this dataset can be used to study structural relationships between different parts of the human body when performing actions and fusion approaches that involve multi-modal sensor data.
\end{abstract}

\begin{IEEEkeywords}
Human action recognition, multi-modal human action dataset, Changzhou University: a comprehensive multi-modal human action dataset, multi-modal feature fusion.
\end{IEEEkeywords}
\end{minipage}}}

\maketitle

\section{Introduction}
\label{sec:introduction}
Human action recognition(HAR) has been applied to intelligent home, safety monitoring, human-computer interaction, health monitoring, rehabilitation training and other fields\cite{Luvizon:2018,Fabien:2018,Hou:2017}. 
The research of human action recognition benefits from the development of sensor technology, such as RGB camera, depth camera, accelerometer and gyroscope\cite{Hou:2017}.

Using video data collected by RGB cameras to recognize human movements has been widely studied, such as Two Stream CNN\cite{Simonyan:2014}, TSN\cite{Wang:2016} and C3D\cite{Tran:2015}.
There are RGB camera-based datasets such as UCF101\cite{Soomro:2012}, HMDB51\cite{Kuehne:2011}, 
Olympic sports Dataset\cite{Niebles:2010}, UT-Interaction\cite{Ryoo:2010} and BEHAVE\cite{Blunsden:2010} that have facilitated the comparison of different recognition approaches.
Depth images have a few advantages over conventional RGB images for human action recognition.  
Depth images are able to provide three-dimensional structure and motion information towards distinguishing different actions\cite{Chen:2015}. 
They are also relatively insensitive to changes in lighting conditions. 
So far, scholars have made many excellent achievements by depth images.
Li \emph{et al}.\cite{Li:2020} presented depth spatial-temporal maps (DSTM), which provide compact global spatial and temporal information.
Chao \emph{et al}.\cite{Chao:20204} proposed depth spatial-temporal energy map (DSTEM) for representing the temporal information, energy information, and spatial structure information of actions.
Yang \emph{et al}.\cite{Yang:2020} proposed a method by Multi-label subspace Learning (MLSL) to fuse depth sequential information entropy map (DSIEM) and skeleton data.
Chao \emph{et al}.\cite{Chao:20209} proposed a motion collaborative Spatio-temporal vector (MCSTV), which comprehensively considers the integral and cooperative between the moving joints of the human body.
There exist human action datasets of depth images that have been created using the kinect camera, e.g.
MSR Action 3D Dataset\cite{W:2010}, MSR Daily Activity 3D Dataset\cite{Wang:2012}, Gaming Dataset(G3D)\cite{V:2016}, UTKinect-Action3D Dataset\cite{Xia:2012}, 
Florence 3D Actions Dataset\cite{Seidenari:2013}, NTU RGB+D\cite{Amir:2016}, NTU RGB+D 120\cite{Jun:2019}.

In recent years, with the rapid development of wireless sensor technology, wearable motion capture systems came into being. Compared with the traditional optical motion capture system, 
using wearable sensors to collect data has many advantages, such as not being affected by the light, site, shelter, portability and so on. 
Therefore, research of human action recognition based on wearable motion capture system is becoming a research hotspot. 
Mo \emph{et al}.\cite{Mo:2019} use multiple wearable sensors fixed in the human body to study the contribution of different parts to complete the action. 
Liu \emph{et al}.\cite{Liu:2016,Liu:2015} present an efficient algorithm to identify temporal patterns among actions and utilize the identified patterns to represent activities for automated recognition. 
Chen \emph{et al}.\cite{C:2014} use a wearable inertial sensor to identify the act of pill intake. 
Yurtman \emph{et al}.\cite{Yurtman:2017} focus on invariance to sensor orientation and develop two alternative transformations to remove the effect of absolute sensor orientation from the raw sensor data. 
There are public accessible datasets for human action recognition using wearable inertial sensors which have been used for comparison of recognition algorithms (e.g., PAMAP2 Dataset\cite{Reiss:2012A, Reiss:2012B}, USC-HAD\cite{Zhang:2012}, OPPORTUNITY Dataset\cite{Chavarriaga:2013}, MHEALTH Dataset\cite{Banos:2014}).

Depth cameras and wearable inertial sensors are mostly used in human action recognition independently.
In other words, the simultaneous utilization of both depth cameras and inertial sensors for action recognition has been restricted in the literature.
Mo \emph{et al}.\cite{Mo:2019} use joint feature selection and subspace learning (JFSSL)\cite{WANG:2016} method to fuse features of different modals to achieve robust action recognition. 
Elmadany \emph{et al}.\cite{Elmadany:2018} propose bimodal hybrid centroid canonical correlation analysis (BHCCCA) and multi-modal hybrid centroid canonical correlation analysis (MHCCCA) to learn the discriminative and informative shared space for fusing different modal data. 
Liu \emph{et al}.\cite{LiuChen:2014a, LiuChen:2014}use the framework of HMM to fuse data from a depth camera and an inertial sensor for robust hand gesture recognition.
Chen \emph{et al}.\cite{Chen:2014improve} develop a fusion approach based on depth and inertial sensors for enhancing human action recognition.
There are public accessible datasets that include both depth and inertial sensor data, such as UTD-MHAD\cite{Chen:2015}, MHAD\cite{Ofli:2013}, UR Fall Detection Dataset\cite{Kepski:2014} and 50 Salads\cite{Stein:2013}.

In order to facilitate the research of multi-modal sensor fusion for human action recognition, this paper provides a multi-modal human action dataset using kinect depth camera and multiple wearable sensors, which is called Changzhou University multi-modal human action dataset (CZU-MHAD). 
Our dataset contains more wearable sensors, which intends to obtain the position data of human bones and joints, as well as three-axis acceleration and three-axis angular velocity data of 10 main motion joints. 
Our dataset provides time synchronous depth video, skeleton joint position, three-axis acceleration and three-axis angular velocity data to describe a complete human action. 

In this paper, we focus on the challenging problem of human action recognition. Our contributions are as follows:
\begin{itemize}
\itemsep=0pt
\item CZU-MHAD is proposed, which consists of 22 actions and three modals temporal synchronized data. 
\item CZU-MHAD collects three-axis acceleration and three-axis angular velocity data of 10 main motion joints through wearable sensors, which can more accurately describe the information of the main motion joints of the human body.
\end{itemize}

This paper is organized as follows. 
In Section 2, the existing datasets are briefly reviewed. 
In Section 3, CZU-MHAD is detailly described. 
The experimental results are presented in Section 4. 
Finally, Section 5 summarizes the paper.

\section{Existing datasets}
\subsection{Berkeley MHAD}
Berkeley MHAD dataset contains 659 data sequences collected by 12 RGB cameras, 
2 Microsoft kinect cameras, 6 wearable acceleration sensors and 4 microphones. 
In the Berkeley MHAD dataset, 12 subjects (7 males and 5 females) performed 11 actions, 
each of which is repeated 5 times. In the process of data collection, 
subjects are informed in advance of the type of action to be performed, 
but there is no specific description of how to perform the action (such as performance style or speed). 
Therefore, the subjects integrated different styles into some movements (such as boxing and throwing). 

\subsection{UR Fall Detection Dataset}
The UR Fall Detection Dataset focuses on human fall detection. 
The UR Fall Detection Dataset collects 60 depth video sequences, 
color video sequences and acceleration data sequences through 2 Microsoft Kinect cameras and 1 wearable acceleration sensor. 
Among them, the kinect camera is installed on the ceiling, 
and the wearable acceleration sensor is fixed on the back spine of the subject. 
The dataset contains two types of actions (fall and not fall). 

\subsection{UTD-MHAD}
The UTD-MHAD dataset collects 861 data sequences through 1 Microsoft kinect camera and 1 wearable inertial sensor. 
In the UTD-MHAD dataset, 8 subjects (4 males and 4 females) perform 27 actions, each of which is repeated 4 times. 
In the process of data acquisition, a kinect camera is fixed on a tripod about 3 meters in front of the subject to ensure that the subject's body appears completely in the depth image. 
And the wearable inertial sensor is fixed on the right wrist or thigh of the subject according to the type of action. 
The UTD-MHAD dataset has the following characteristics: 
(1) subjects perform the same action at different speeds in many times; 
(2) physiological conditions such as height and weight of subjects are different; 
(3) subjects perform the same action in a natural way.  
For example, in different experiments, the number of times the subjects clapped their hands is different. 

\section{CZU MULTIMODAL HUMAN ACTION DATASET}
\begin{figure}[htb]
\centering
\subfigure[]{
    \includegraphics[height=5cm]{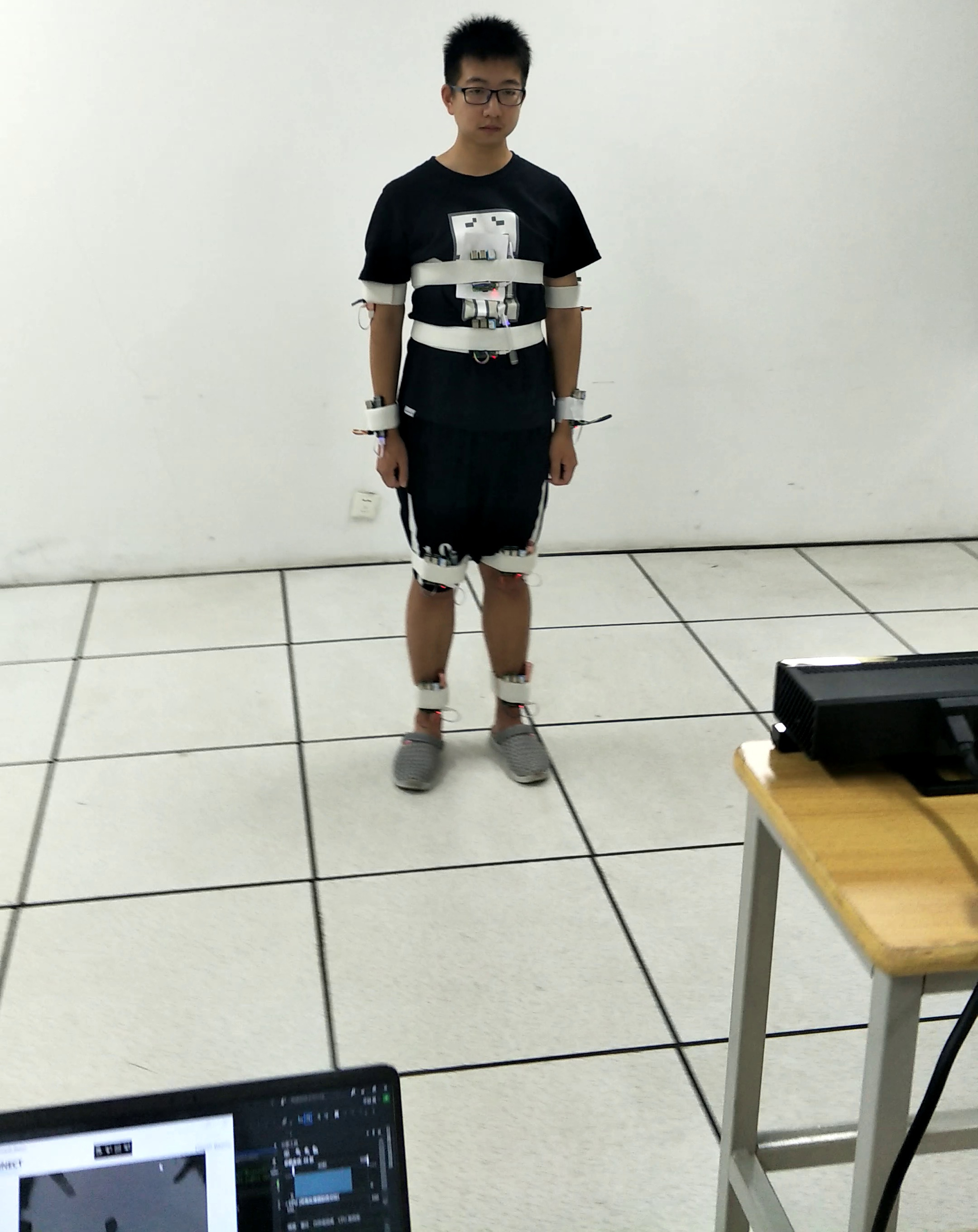}
    \label{Fig1-1}
}
\subfigure[]{
    \includegraphics[height=5cm]{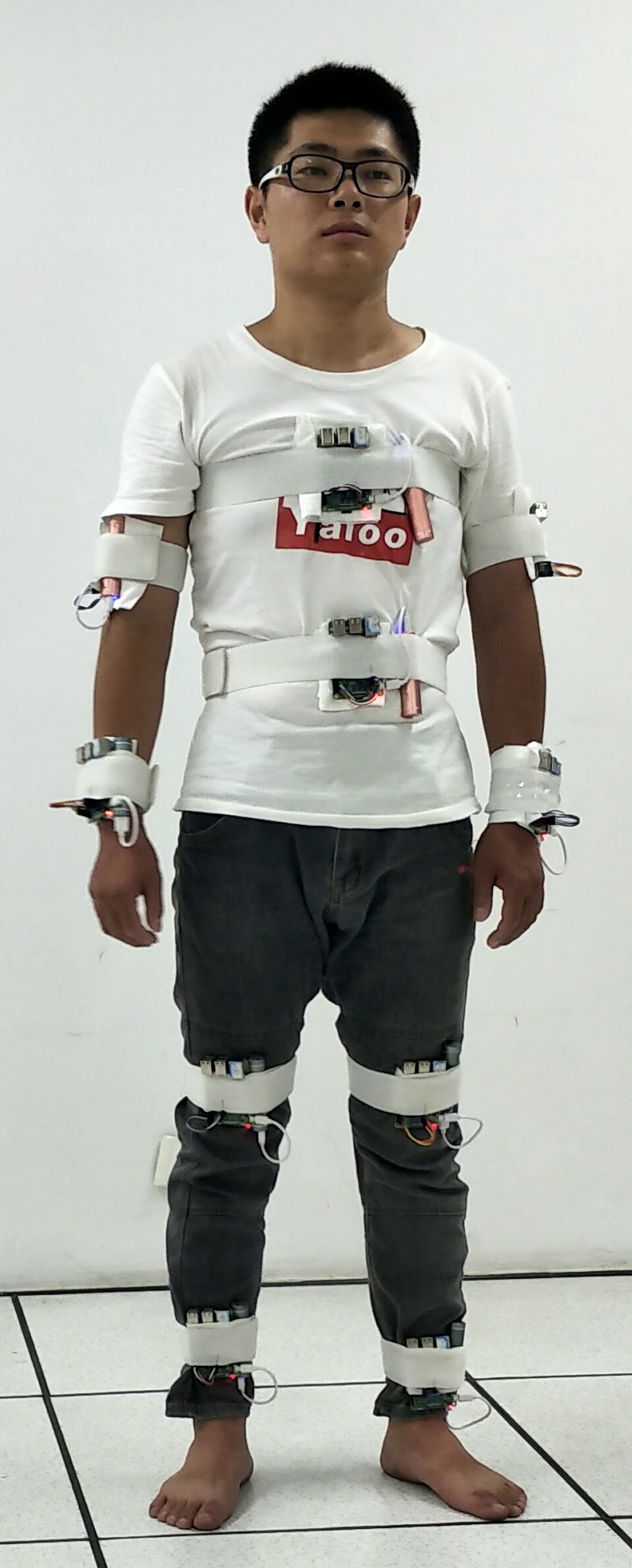}
    \label{Fig1-2}
}
\subfigure[]{
    \includegraphics[height=5cm]{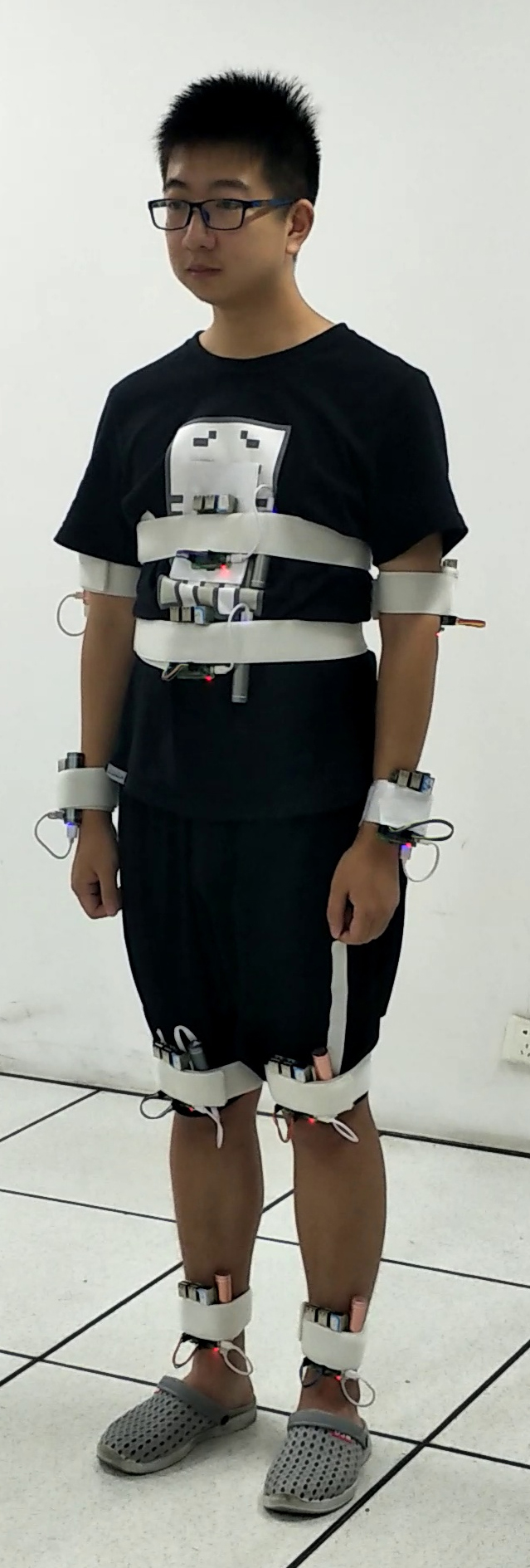}
    \label{Fig1-3}
}
\caption{Real scene. (a) Data acquisition system architecture of the real scene. (b) Subject 1;(c) Subject 2.}
\label{Fig1}
\end{figure}

The CZU-MHAD contains 880 action sequences of 22 different actions collected by 1 Microsoft kinect camera and 10 wearable sensors. 
The real scene is shown in Fig. \ref{Fig1}, Fig. \ref{Fig1-1} shows the data acquisition system architecture of the real scene, Fig. \ref{Fig1-2} shows subject 1, Fig. \ref{Fig1-3} shows subject 2.

\subsection{Sensors}
\begin{figure}[htb]
    \centering
    \includegraphics[width=5cm]{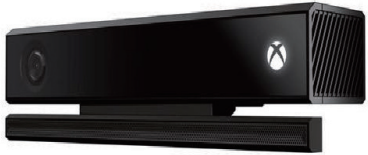}
    \caption{Microsoft kinect camera}
    \label{kinect}
\end{figure}

\begin{figure}[htb]
    \centering
    \includegraphics[width=4.5cm]{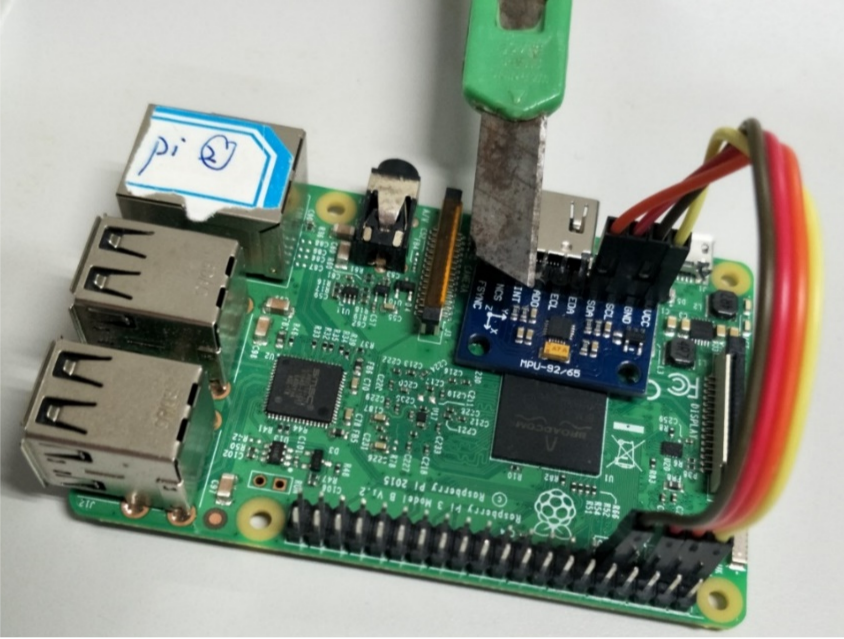}
    \caption{MPU9250}
    \label{MPU9250}
\end{figure}

The CZU-MHAD uses 1 Microsoft kinect camera and 10 wearable sensors. These two kinds of sensors are widely used, which have the characteristics of low power consumption, low cost and simple operation. 
In addition, it does not require too much computing power to process the data collected by the two kinds of sensors in real time. 
Fig. \ref{kinect} shows the Microsoft kinect v2 camera, which can collect both color and depth images at a sampling frequency of 30 frames per second. 
Kinect SDK is a software package provided by Microsoft, which can be used to track 25 skeleton joint points and their 3D spatial positions. 
The MPU9250 can capture 3-axis acceleration, 3-axis angular velocity and 3-axis magnetic intensity. 
Measurement range of MPU9250: the measurement range of accelerometer is $\pm$16g, and the measurement range of angular velocity of the gyroscope is $\pm$2000 degrees/second. 
CZU-MHAD uses Raspberry PI to interact with MPU9250 through the integrated circuit bus (IIC) interface, realizing the functions of reading, saving and uploading MPU9250 sensor data to the server. 
The connection between Raspberry PI and MPU9250 is shown in Fig. \ref{MPU9250}. 

\subsection{Data Acquisition System Architecture}
\begin{figure}[htb]
    \centering
    \includegraphics[width=8.5cm]{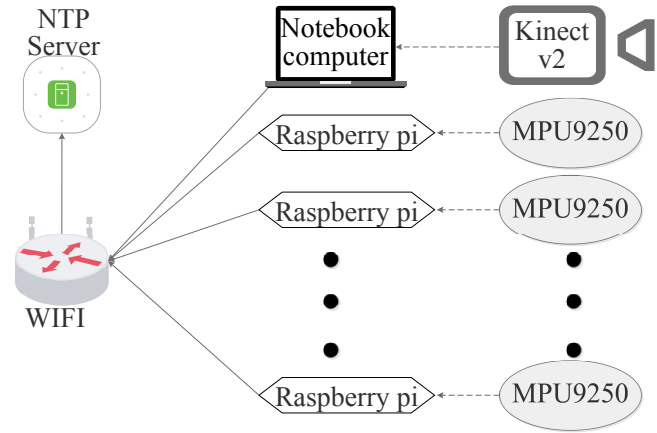}
    \caption{Sampling system architecture.}
    \label{Samplingsystemarchitecture}
\end{figure}

The CZU-MHAD uses a kinect v2 camera to collect depth image and skeleton joint position data, and uses 10 MPU9250 sensors to collect 3-axis acceleration and 3-axis angular velocity data.
In order to collect the 3-axis acceleration data and the 3-axis angular velocity data of the whole body, a motion data acquisition system including 10 MPU9250 sensors is built-in this paper. 
The MPU9250 sensor is controlled by Raspberry PI and the kinect v2 is controlled by a notebook computer,
time synchronization with a NTP server is carried out every time data is collected.
The sampling system architecture is shown in Fig. \ref{Samplingsystemarchitecture}.
    
\begin{figure}[htb]
    \centering
    \includegraphics[width=2in]{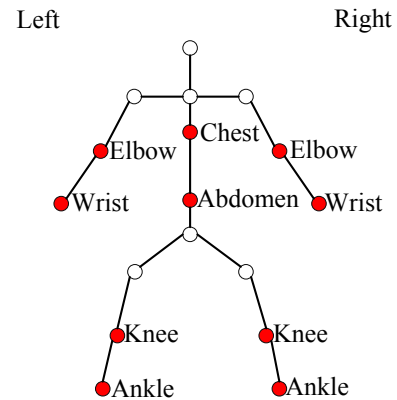}
    \caption{Position Schematic Diagram Of Wearable Sensor.}
    \label{Positionschematicdiagramofwearablesensor}
\end{figure}

The position of wearable sensors is determined as shown in Fig. \ref{Positionschematicdiagramofwearablesensor}, and the points marked are the positions of inertial sensors. 
The Left in the figure is the left side of the human body, and the Right is the right side of the human body. 
Due to wrists, elbows, knees and ankles are at the junction of human bones, the movement of these joints is most obvious when the limbs are moving. 
Chest and Abdomen are located in the trunk of the body, the motion information of these joints is particularly important when recognizing turning and twisting. 
Considering 3-axis acceleration and 3-axis angular velocity information of these joints comprehensively, most actions can be recognized effectively. 
Especially in recognizing actions with small differences, the accuracy can be greatly improved.

\subsection{Background removal of depth image}
In CZU-MHAD, the method of depth motion map (DMM)\cite{Chen:2016RealTime} is used to project the depth images into three orthogonal cartesian planes to realize action recognition. 
Before depth image projection, we need to extract the foreground of the moving human body. 
In order to make the calculation of this task more efficient, 
in this paper, the subjects are placed in the designated position during the processing of data collection.

\begin{figure}[htb]
    \centering
    \subfigure[]{
        \includegraphics[width=4cm]{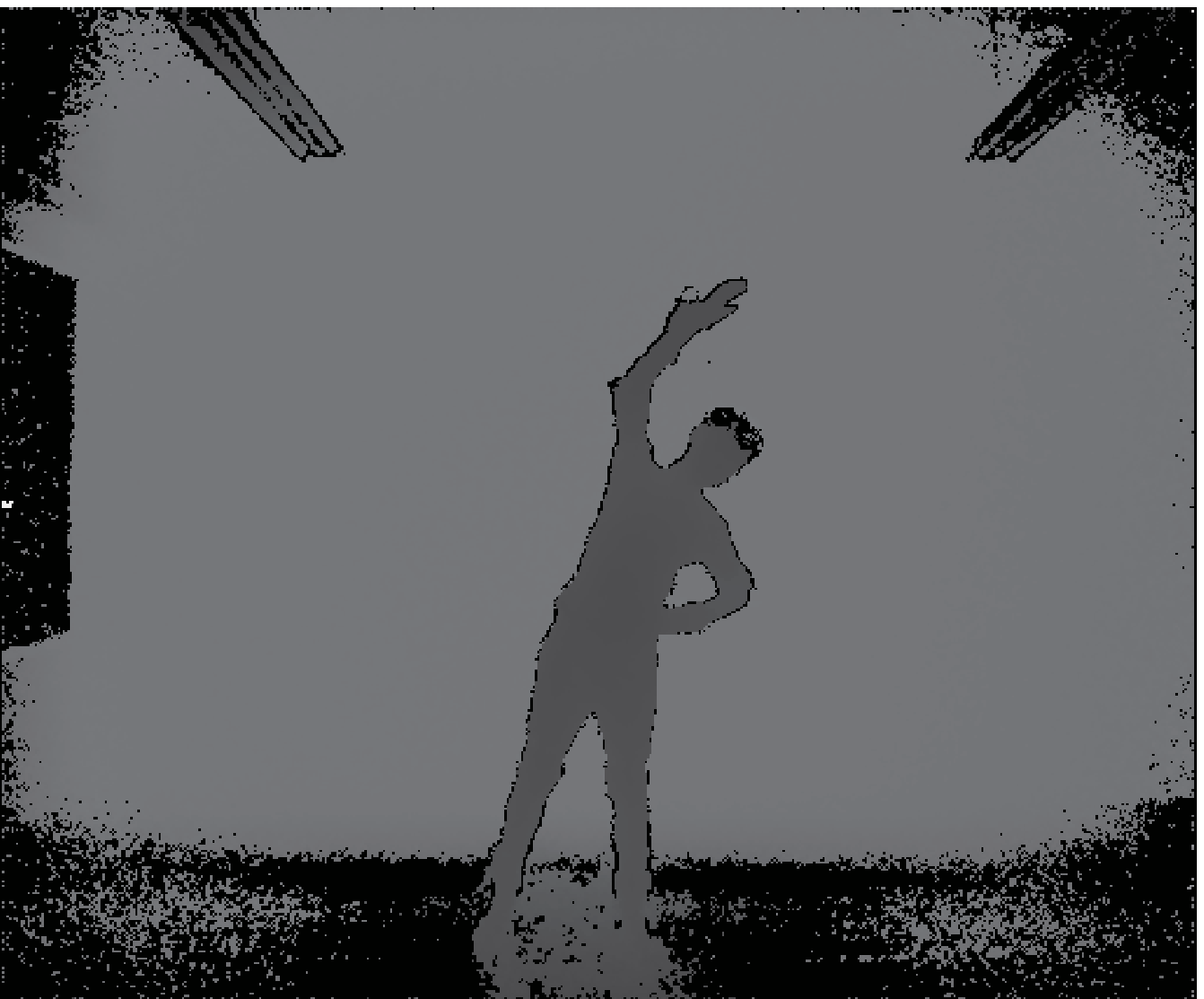}
        \label{removeback1}
    }
    \subfigure[]{
        \includegraphics[width=4cm]{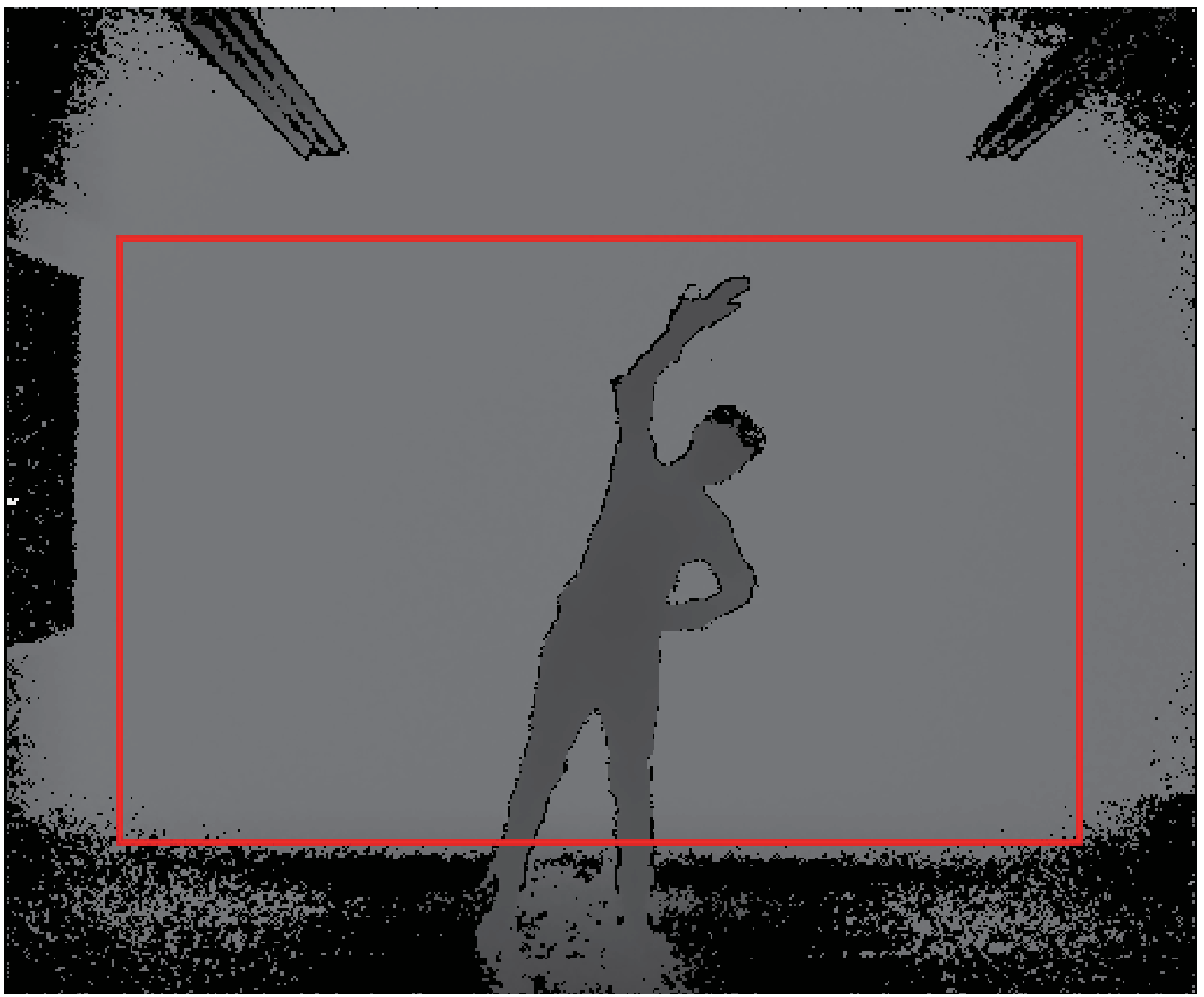}
        \label{removeback2}
    }
    \subfigure[]{
        \includegraphics[width=4cm]{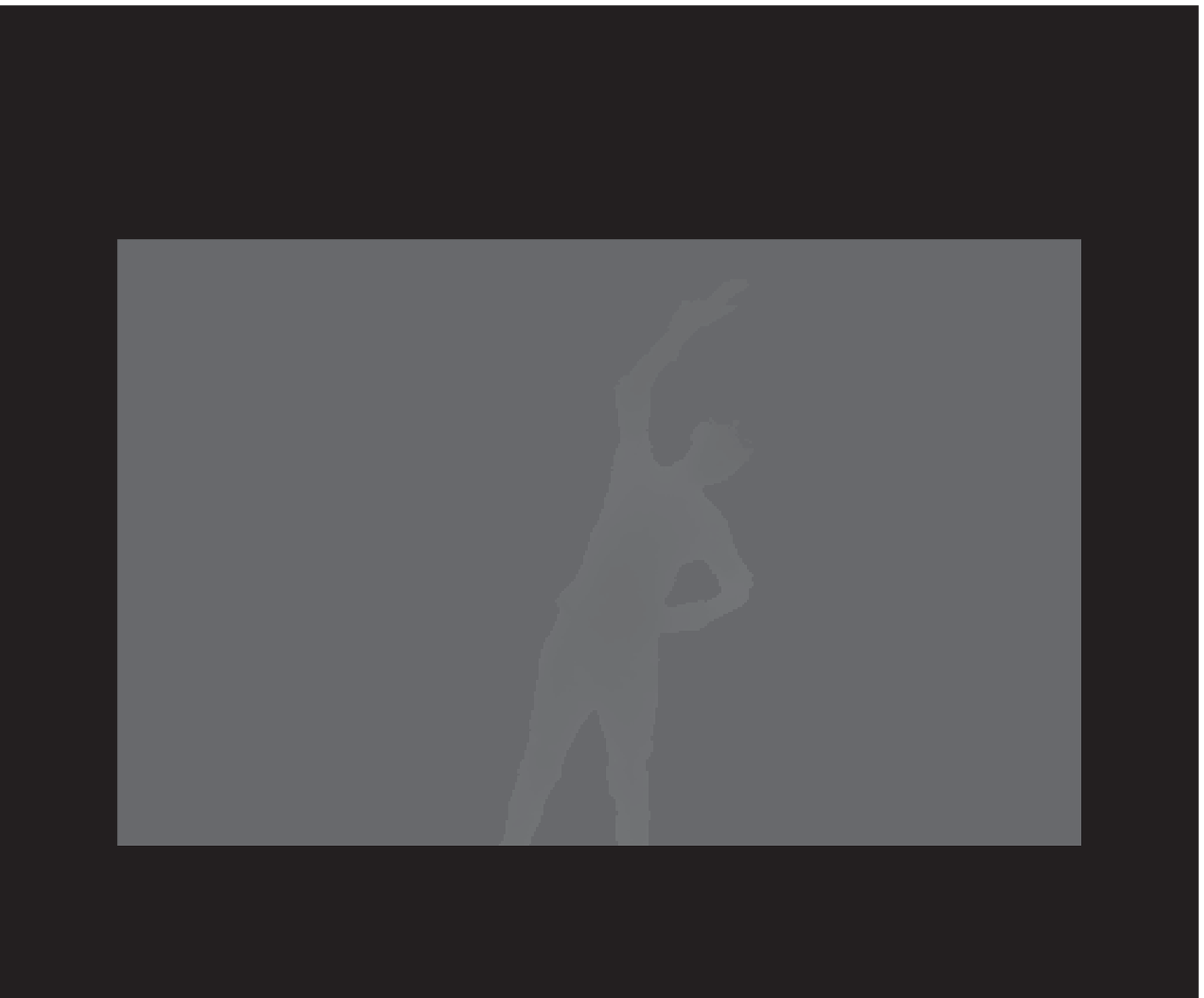}
        \label{removeback3}
    }
    \subfigure[]{
        \includegraphics[width=4cm]{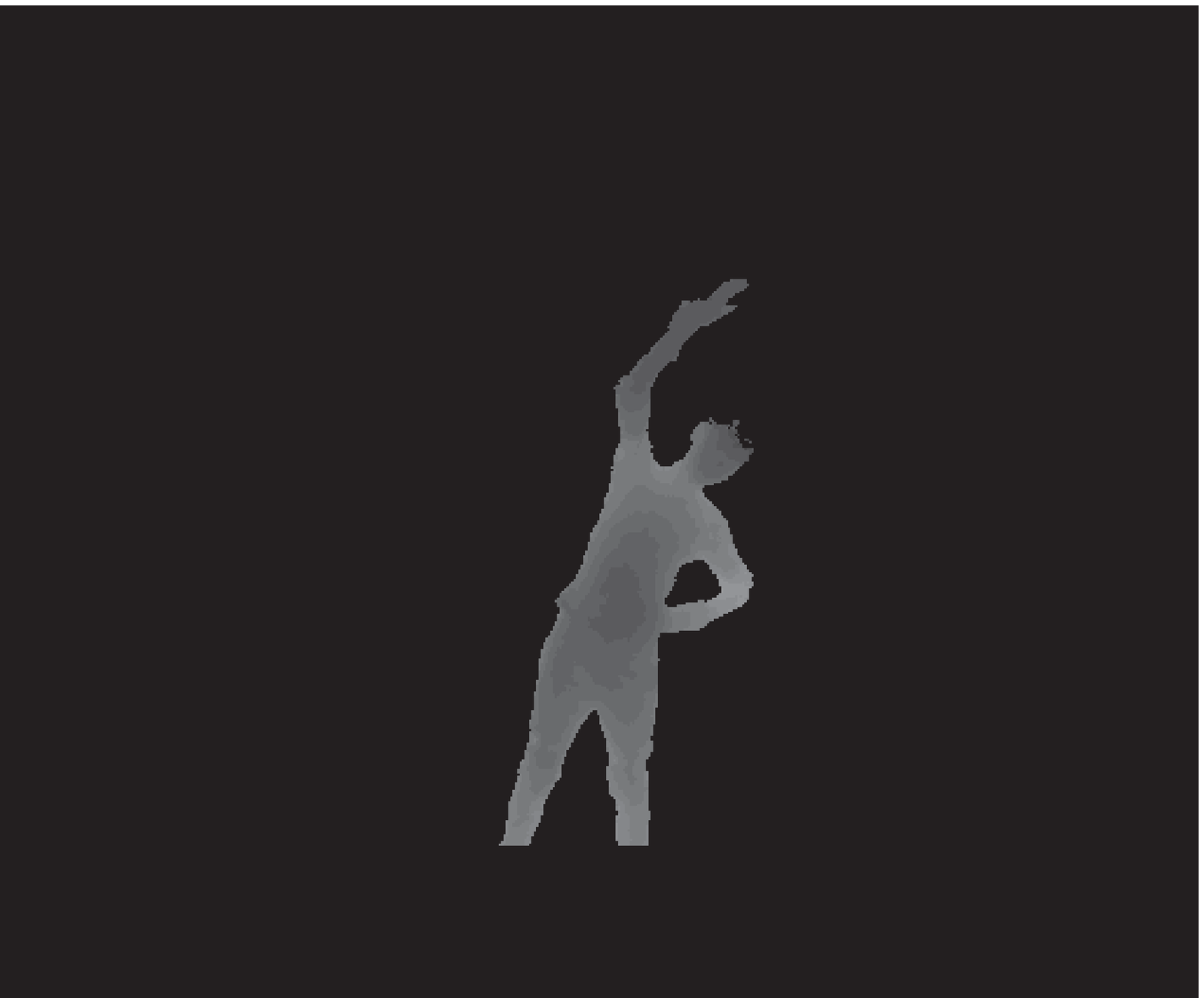}
        \label{removeback4}
    }
    \caption{The process of extracting the foreground of moving human body.
    (a) Step 1;
    (b) Step 2;
    (c) Step 3;
    (d) Step 4.}
    \label{removebackground}
\end{figure}

In the first step of extracting the foreground of moving human body, 
we determine the largest rectangular box including human body and excluding other unrelated objects, the rectangular box is shown in Fig. \ref{removeback2}.
Then we set the gray value of the area outside the rectangular box to 0 ,as shown in Fig. \ref{removeback3}. 
Because the rectangular box contains foreground and background, and the gray value distribution of foreground and background is different. 

\begin{figure}[htb]
    \centering
    \includegraphics[width=8.5cm]{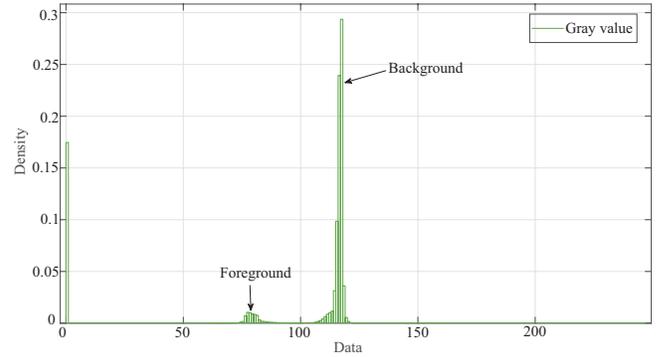}
    \caption{The probability distribution of gray value of depth image.}
    \label{grayvalue}
\end{figure}

In order to extract the foreground from the background, this paper makes histogram statistics on the gray value of the depth image, it is shown in Fig. \ref{grayvalue}.
According to Fig. \ref{grayvalue}, we can know the gray value range of subjects and background in depth image, 
and finally determine the gray value range of foreground including human body as [70,98]. 
Finally, the contrast of the depth image is enhanced by the piecewise linear function \ref{equationgrayvalue}. 
The results of the extracted human motion foreground are shown in the Fig. \ref{removeback4}.

\begin{equation}
    f(x)=\left\{\begin{array}{l}{0, x<70} \\ {\frac{x-70}{28} * 256,70 \leq x \leq 98} \\ {0, x>98}\end{array}\right.
    \label{equationgrayvalue}
\end{equation}
Where $x$ is the gray value of the original depth image, $f(x)$ is the gray value of the depth image after data processing.

\subsection{Dataset description}
The CZU-MHAD dataset contains 22 actions performed by 5 subjects (5 males). 
Each subject repeats each action 8 times. 
The CZU-MHAD dataset contains a total of 880 samples. 
The 22 actions performed are listed in Table. \ref{tableexample}.
It can be seen that CZU-MHAD includes common gestures (such as Draw $x$, Draw circle), 
daily activities (such as Sur Place, Clap, Bend down), 
and training actions (such as Left body turning movement, Left lateral movement). 
The CZU-MHAD dataset can be downloaded by visiting the link https://github.com/yujmo/czu\_mhad/.

\begin{table}[htb]
\caption{Human actions in CZU-MHAD}
\label{tableexample}
\centering
\begin{tabular}{|p{50pt}|p{150pt}|}  
\hline
ID & Action name \\  
\hline
1 & Right high wave  \\
2 & Left high wave   \\
3 & Right horizontal wave  \\
4 & Left horizontal wave   \\
5 & Hammer with right hand  \\
6 & Grasp with right hand  \\
7 & Draw $x$ with right hand  \\
8 & Draw $x$ with left hand    \\
9 & Draw circle with right hand  \\
10 & Draw circle with left hand   \\
11 & Right foot kick forward  \\
12 & Left foot kick forward  \\
13 & Right foot kick side  \\
14 & Left foot kick side   \\
15 & Clap                \\
16 & Bend down       \\
17 & Wave up and down             \\
18 & Sur Place                    \\
19 & Left body turning movement   \\
20 & Right body turning movement  \\
21 & Left lateral movement        \\
22 & Right lateral movement       \\
\hline 
\end{tabular}
\end{table}

The comparison between the self-built dataset and the public datasets (MHAD, WARD, UTD-MHA, MSR-Action3D, MSRDailyActivity3D, NTU RGB+D, NTU RGB+D 120, etc.) is shown in Table. \ref{table2}.

\begin{table}[t]
    \caption{Comparison of multiple databases.}
    \label{table2}
    \setlength{\tabcolsep}{3pt}
    \begin{tabular}{|p{85pt}|p{24pt}|p{24pt}|p{24pt}|p{30pt}|p{25pt}|}
    \hline
    Database & Sensors & Subject & Action & Sequence & Time\\
    \hline
    MHAD& 6& 12& 11& \textgreater{647}& 5\\
    WARD& 5& 20& 13& 1300& 5\\
    UTD-MHAD& 1& 8& 27& 861& 4\\
    MSR-Action3D& 0& 10& 20& 567& 2 or 3\\
    MSR Daily Activity 3D& 0& 10& 16& 320& 2\\
    NTU RGB+D& 0& 40& 60& 56880& \textgreater{20}\\
    NTU RGB+D 120& 0& 106& 120& 114480& 9\\
    {\color{red} \textbf{CZU-MHAD}(ours)}& {\color{red} \textbf{10}}& {\color{red} \textbf{7}}& {\color{red} \textbf{22}}& {\color{red} \textbf{\textgreater{1540}}}& {\color{red} \textbf{10}}\\
    \hline
    \end{tabular}
\end{table}

It can be seen from the Table. \ref{table2} that the number of sensors in the public datasets is little and cannot fully represent human motion information.
CZU-MHAD has 10 wearable sensors, it can more accurately describe the motion of several major joints when the human body is active.

\begin{figure*}[htb]
    \centering
    \includegraphics[width=18cm]{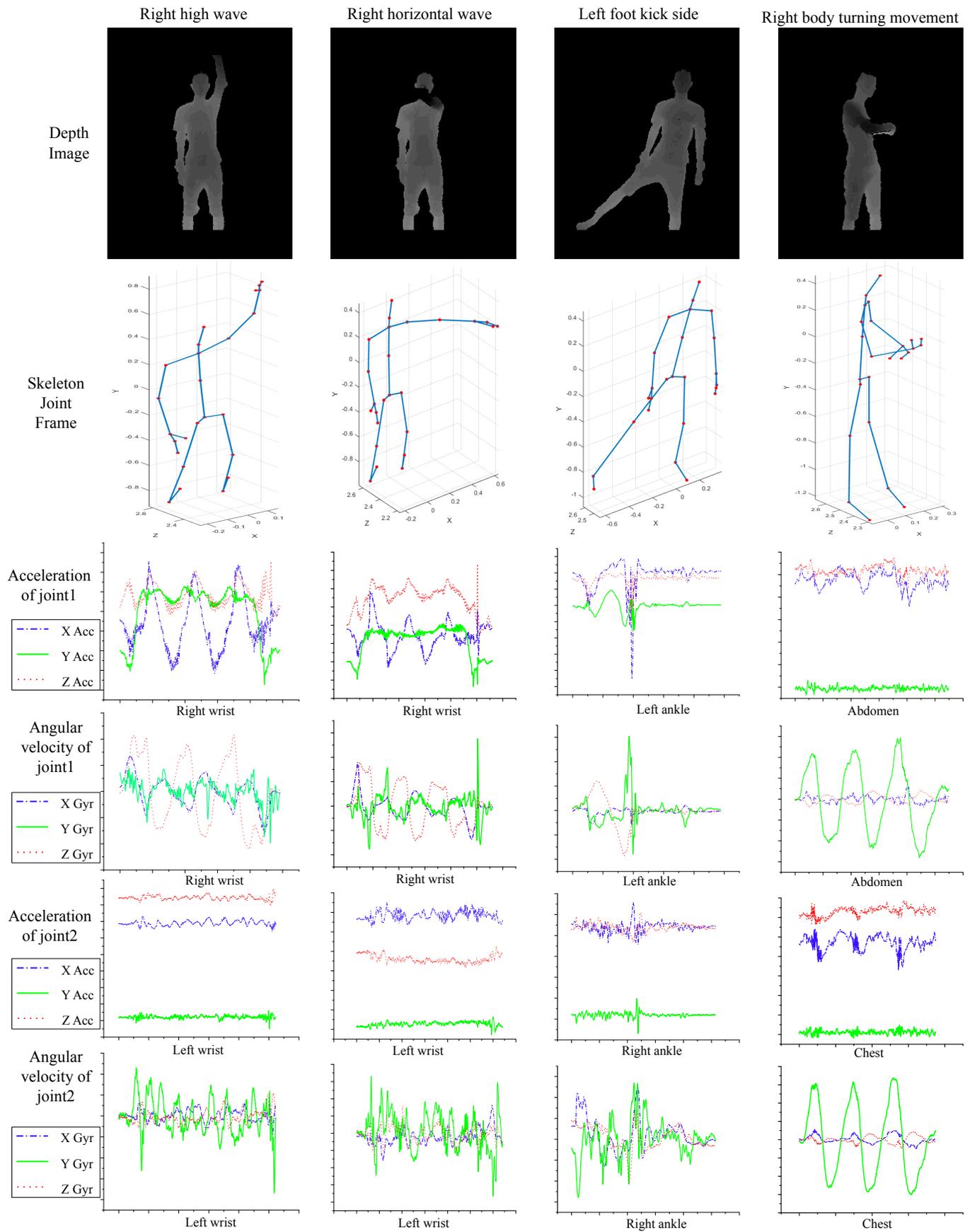}
    \caption{Various modals data of right high wave, right horizontal wave, left foot kick side and right body turning movement.}
    \label{data}
\end{figure*}

Fig. \ref{data} shows various modals data of right high wave, right horizontal wave, left foot kick side and right body turning movement. 
These modals include depth video sequences, skeleton joint coordinates, 3-axis acceleration and 3-axis angular velocity data.
For the right high wave and the right horizontal wave, the main motion area is the right upper body, and the other areas are additional motion areas. 
As can be seen from the figure, the change of the 3-axis acceleration and 3-axis angular velocity of the right wrist is more intense than that of the left wrist.
In addition, it also can be seen that the axis with the largest acceleration and angular velocity change of right high wave and right horizontal wave is different.
The reason is that the direction of the right wrist movement is different.
For the left foot kick side, the main motion area is the left lower body, and the other areas are additional motion areas. 
As can be seen from the figure, the change of the 3-axis acceleration and 3-axis angular velocity of the left ankle is more evident than that of the right ankle.
For the right body turning movement, the main motion area is trunk, and the other areas are additional motion areas. 
As can be seen from the figure, the change of the 3-axis acceleration and 3-axis angular velocity of the abdomen and chest are obvious.

When the human body performs different actions, the contribution degree of different parts is different.
Mo \emph{et al}.\cite{Mo:2019} use multiple wearable sensors to study the structural relationship of different parts when the human body performs different actions.
In addition, CZU-MHAD can be used for human action recognition, posture estimation, disease diagnosis (such as stroke, Spinal Cord Injury (SCI), Parkinson’s Disease (PD) or patients with other physical impairments.), 
rehabilitation treatment, motion detection and analysis, and other fields. Especially in terms of action coordination, 
3-axis acceleration and 3-axis angular velocity collected by 10 wearable sensors in this dataset can be used for subsequent researchers to make more choices for different tasks.

\section{Experiments}
To demonstrate the utility of this multimodality dataset for human action recognition, 
this section provides the outcome of a data fusion approach for human action recognition when using our CZU-MHAD dataset.

\subsection{Experimental setup}
There are two experimental settings in this paper.

\emph{Set 1}: In the first group of experiments, 3/8 samples are used as training samples and the remaining samples as testing samples; 
in the second group, 4/8 samples are used as training samples; 
in the third group, 5/8 samples are used as training samples; 
and in the fourth group, 6/8 samples are used as training samples.
T1$\sim$T4 are used to represent the above four groups of experiments. 

\emph{Set 2}: In the first group of experiments, the data marked as object 1 and 2 are used as training samples and the remaining samples as testing samples; 
in the second group, the data marked as object 1, 2 and 5 are used as training samples; 
in the third group, the data marked as object 1, 2, 3 and 5 are used as training samples. 
T5$\sim$T7 are used to represent the above three groups of experiments.

For action recognition, we use the collaborative representation classifier(CRC) as classifier. 
The regularization parameter $\lambda$ of the CRC classifier is 0.0001.

\subsection{Human action recognition based on 3-axis acceleration}
In this section, we use the 3-axis acceleration of 10 wearable sensors 
to calculate the contribution degree of different parts of the human body to the completion of the action 
and transform it into a structural feature model of cooperative motion \cite{Mo:2019}. 
Then, the model is used to constrain the features of different parts of the human body unsupervised and adaptively. 
We extract the mean, variance, standard deviation, kurtosis and skewness value of the 3-axis acceleration of 
10 wearable sensors as the motion features for human action recognition.

\begin{figure}[htb]
\centering
\includegraphics[width=8.5cm]{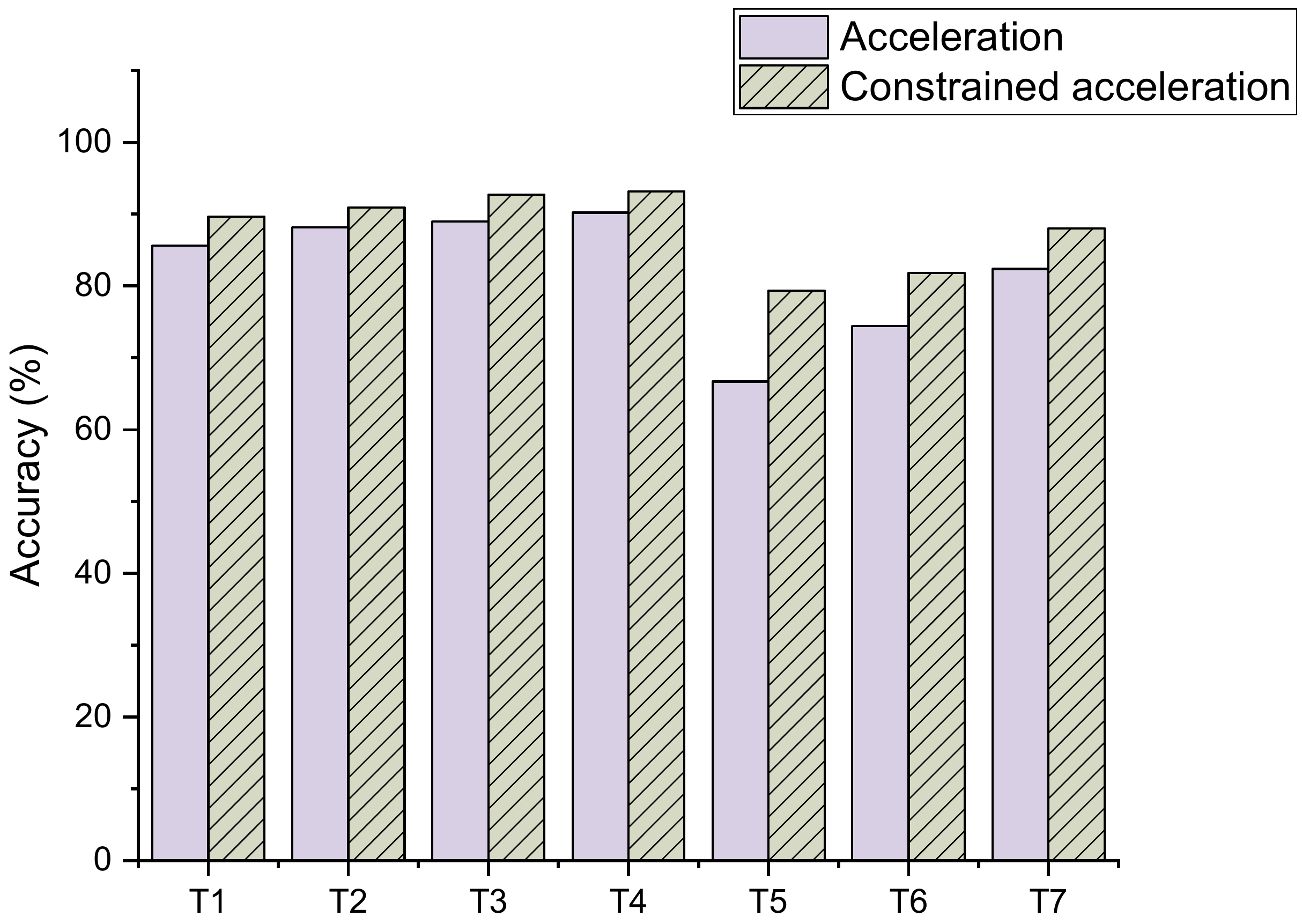}
\caption{Recognition rate of triaxial acceleration and constrained triaxial acceleration.}
\label{Constrained}
\end{figure}

We extract the features of the original triaxial acceleration and the constrained triaxial acceleration respectively, and use them to identify the actions.
The results are shown in the Fig. \ref{Constrained}.
It can be seen that in the T1$\sim$T7 experiment, using the structural feature model of cooperative motion to constrain the features of different parts can effectively 
improve the recognition accuracy of actions. 
And the recognition accuracy of closed test T1$\sim$T4 is higher than that of open test T5$\sim$T7.

\subsection{Human action recognition based on depth image}
In this section, we extract DMM, DSTM, DSTEM, MSTM from the depth video sequence,
and then extract Histograms of Oriented Gradients (HOG) \cite{Yang:2012ACM}, Local Binary Pattern (LBP) \cite{Chen:2015WACV} and Gradient Local Auto-Correlations (GLAC) \cite{Chen:2015ISVC} from DMM,
and extract HOG and LBP from DSTM, DSTEM, and MSTM respectively.
Next, we calculate the recognition rate of DMM-HOG, DMM-LBP, DMM-GLAC, DSTM-HOG, DSTM-LBP, DSTEM-HOG, DSTEM-LBP, MSTM-HOG, and MSTM-LBP,
and the results are shown in Table. \ref{depth image}.

\begin{table}[htbp]  
\caption{Recognition rate (\%) of different methods.}
\label{depth image}
\begin{tabular}{|p{48pt}|p{15pt}|p{15pt}|p{15pt}|p{15pt}|p{15pt}|p{15pt}|p{15pt}|}  
\hline
Method & T1 & T2 & T3 & T4 & T5 & T6 & T7\\  
\hline
DMM-HOG & 96.73 & 97.27 & 97.58 & 96.82 & 88.26 & 88.07 & 90.34    \\
DMM-LBP & 97.09 & 96.36 & 96.06 & 96.34 & 84.28 & 90.34 & 91.48    \\
DMM-GLAC & 93.87 & 95.00 & 96.89 & 98.67 & 80.56 & 91.25 & 92.50    \\
DSTM-HOG & 84.00& 87.50& 90.61& 91.82& 60.98& 68.75& 82.39\\
DSTM-LBP & 83.63& 87.27& 91.52& 93.64& 59.28& 67.90& 81.82\\
DSTEM-HOG & 84.36& 87.50& 88.48& 92.27& 63.26& 71.02& 82.95\\
DSTEM-LBP & 84.91& 87.27& 90.91& 91.82& 63.07& 72.44& 74.43\\
MSTM-HOG & 79.57& 83.21& 84.47& 87.86& 58.81& 66.15& 81.60\\
MSTM-LBP & 82.91& 86.82& 89.09& 93.64& 57.20& 61.36& 75.00\\
\hline 
\end{tabular}
\end{table}

It can be seen from Table. \ref{depth image} that in the T1$\sim$T7 experiment, DMM-HOG, DMM-LBP and DMM-GLAC all maintain high recognition accuracy. 
Especially, DMM-LBP has high and stable recognition rate in all tests, without great fluctuation.
DSTM, DSTEM, and MSTM maintain high recognition accuracy in some experiments, 
but the recognition rate is lower than DMM in some experiments.
The reason is that DSTM, DSTEM, and MSTM focus on expressing the temporal information of actions, 
but in this experiment, the focus is on the spatial structure of actions.

\subsection{Human action recognition based on skeleton}
\begin{figure}[htb]
    \centering
    \includegraphics[width=8.5cm]{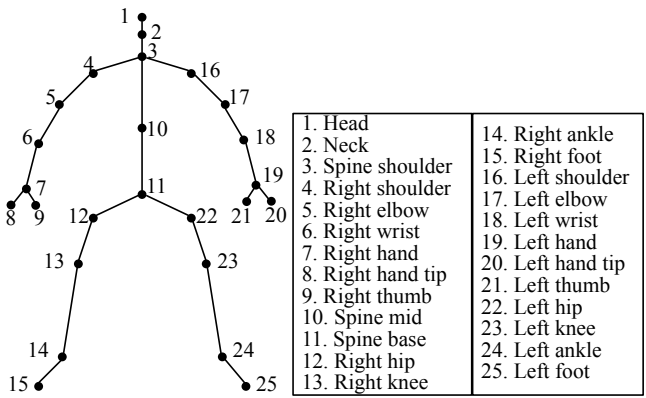}
    \caption{Human skeleton.}
    \label{Skeleton_image}
\end{figure}

The 3-axis position data of 25 human skeleton joints are extracted from the depth images by kinect v2 SDK.
First, the human body is extracted from the complex background using a separation strategy.
Next, the images are evaluated to identify different parts of the body.
Then, machine learning algorithms such as decision tree classifier are used to find the skeleton joints of each part of the body.
Finally, the skeleton system is generated by tracking 25 human skeleton joints.
Human skeleton extracted by kinect v2 is shown in Fig. \ref{Skeleton_image}.
Human skeleton joint frame is shown in Fig. \ref{data}.

\begin{figure}[htb]
\centering
\includegraphics[width=8.5cm]{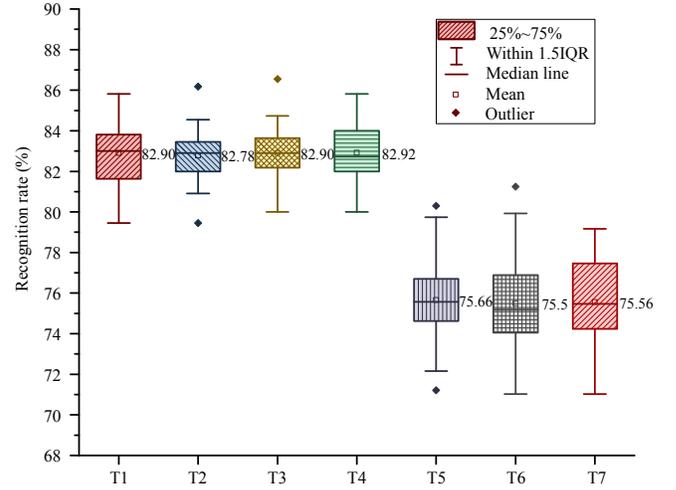}
\caption{Recognition rate (\%) of Skeleton.}
\label{Recognition rate of Skeleton}
\end{figure}

Similarly, we extract mean, variance, standard deviation, kurtosis and skewness value of the 3-axis joint position data of 25 joints as the motion features for human action recognition,
and calculate the recognition rate of skeleton feature in the T1$\sim$T7 experiment, and the results are shown in Fig. \ref{Recognition rate of Skeleton}.

The data in Fig. \ref{Recognition rate of Skeleton} is the recognition rates of skeleton feature running 50 times.
The figure clearly shows the mean, maximum, minimum, median, outlier, upper quartile, and lower quartile of the multiple results.
It can be seen from Fig. \ref{Recognition rate of Skeleton} that in the T1$\sim$T7 experiment, skeleton feature maintain high and stable recognition accuracy, without great fluctuation. 

\subsection{Multi-modal feature fusion}
JFSSL projects multi-modal features into subspace for cross-modal content retrieval, and the method constrains projection matrix based on $\ell_{21}$ norm.
The minimization problems are as follows:
\begin{equation}
    \begin{split}
        \min _{\mathbf{U}_{1}, \ldots, \mathbf{U}_{M}} \sum_{p=1}^{M}\left\|\mathbf{X}_{p}^{T} \mathbf{U}_{p}-\mathbf{Y}\right\|_{F}^{2}+\lambda_{1} \sum_{p=1}^{M}\left\|\mathbf{U}_{p}\right\|_{21}\\
        +\lambda_{2} \Omega\left(\mathbf{U}_{1}, \ldots , \mathbf{U}_{M}\right)
    \end{split}
    \label{JFSSL1}
\end{equation}

Where $\mathrm{U}_{p}, p=1,2, \ldots, M$ is the projection matrix of the $p$-th modality. 
$\mathrm{X}_{p}$ is the sample features of the $p$-th modality before the projection. 
$\mathrm{X}_{p}^{T} \mathrm{U}_{p}$ is the sample features of the p-th modality after the projection. 
Y is the primary target projection matrix in the subspace, $\mathrm{Y}=\left\{y_{1}, y_{2}, \cdots, y_{N}\right\}^{T}$.
$\lambda_{1}$ and $\lambda_{2}$ are weighting parameters.

In Eq. \ref{JFSSL1}, the first term is used to learn projection matrix, 
the second term is used for feature selection, 
and the third term is used to maintain similarity between modals.

Eq. \ref{JFSSL1} is derived and the iterative formula of projection matrix is obtained:
\begin{equation}
    \begin{split}
        \mathbf{U}_{p}^{t+1}=&\left(\mathbf{X}_{p} \mathbf{X}_{p}^{T}+\lambda_{1} \mathbf{R}_{p}+\lambda_{2} \mathbf{X}_{p} \mathbf{L}_{p p}\left(\mathbf{X}_{p}\right)^{T}\right)^{-1}\\
        &\left(\mathbf{X}_{p} \mathbf{Y}-\lambda_{2} \sum_{q \neq p} \mathbf{X}_{p} \mathbf{L}_{p q}\left(\mathbf{X}_{q}\right)^{T} \mathbf{U}_{q}^{t}\right)
        \label{JFSSL2}
    \end{split}
\end{equation}
Where $\mathbf{R}_{p}=\operatorname{Diag}\left(r_{p}\right), r_{p}^{i}=\frac{1}{2 \left\|u_{p}^{i}\right\|}$.
$\mathbf{L} = \mathbf{D}-\mathbf{W}$ is Laplace matrix.

In this paper, the multimode projection matrix is obtained by iteration Eq. \ref{JFSSL2}. 
Then action recognition is carried out according to the fusion feature by Eq. \ref{JFSSL1}.

In this section, we use a cross-media retrieval method JFSSL to carry out feature selection and multi-modal feature fusion. 
JFSSL is used to fuse 3-axis acceleration feature, skeleton joint point position feature and depth image feature,
and realize the feature selection in the process of feature fusion.

\begin{figure}[htb]
\centering
\includegraphics[width=8.5cm]{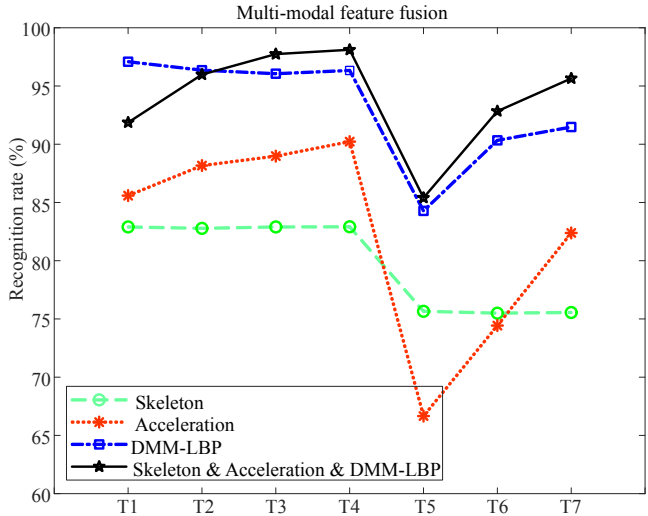}
\caption{Recognition rate of multi-modal feature fusion.}
\label{JFSSL}
\end{figure}

The recognition performance of the feature fusion is compared with the performance of each individual modal feature. 
The results obtained are displayed in Fig. \ref{JFSSL}.
It can be seen that the fusion of multi-modal features can effectively improve the recognition accuracy.

\section{Conclusion}
This paper presents a public dataset called CZU-MHAD, which contains 10 wearable sensors, effectively solves the shortcomings in the layout and number of sensors,
can fully represent the action features. The dataset includes three data modals, including depth video, skeleton position and inertia signals (acceleration and rotation signals). 
It contains 880 data sequences of 22 human actions performed by 5 subjects. This public dataset helps research groups to carry out various forms of research activities for human action recognition. 


\end{document}